\documentclass[runningheads]{llncs}

\usepackage{eccv}

\usepackage{eccvabbrv}

\usepackage{graphicx}
\usepackage{booktabs}

\usepackage[accsupp]{axessibility}  

\usepackage{hyperref}

\usepackage{orcidlink}

\begin{document}

\title{How Knowledge Distillation Mitigates the Synthetic Gap in Fair Face Recognition} 

\titlerunning{KD to Mitigate the Synthetic Gap}

\author{Pedro C. Neto\inst{1,2,4}\orcidlink{0000-0003-1333-4889} \and
Ivona Colakovic\inst{3,4}\orcidlink{0000-0002-1603-9458} \and
Sašo Karakatič\inst{3}\orcidlink{0000-0003-4441-9690}
\and
Ana F. Sequeira\inst{1,2}\orcidlink{0000-0002-6685-2033}}

\authorrunning{Pedro C. Neto et al.}

\institute{INESC TEC, Porto, Portugal \and
Faculty of Engineering of the University of Porto, Porto, Portugal \\
\email{\{pedro.d.carneiro,ana.f.sequeira\}@inesctec.pt}\\\and
Faculty of Electrical Engineering and Computer Science, University of Maribor, Maribor, Slovenia\\\and
These authors contributed equally\\
\email{\{ivona.colakovic,saso.karakatic\}@um.si}}

\maketitle

\begin{abstract}

 Leveraging the capabilities of Knowledge Distillation (KD) strategies, we devise a strategy to fight the recent retraction of face recognition datasets. Given a pretrained Teacher model trained on a real dataset, we show that carefully utilising synthetic datasets, or a mix between real and synthetic datasets to distil knowledge from this teacher to smaller students can yield surprising results. In this sense, we trained 33 different models with and without KD, on different datasets, with different architectures and losses. And our findings are consistent, using KD leads to performance gains across all ethnicities and decreased bias. In addition, it helps to mitigate the performance gap between real and synthetic datasets. This approach addresses the limitations of synthetic data training, improving both the accuracy and fairness of face recognition models.

\keywords{Knowledge distillation \and Fairness \and Face recognition \and Synthetic data}
\end{abstract}

\section{Introduction}

Face recognition (FR) datasets have been one of the strongest elements to fuel the evolution of FR systems. From general datasets with millions of images~\cite{DBLP:conf/cvpr/DengGXZ19,DBLP:conf/eccv/GuoZHHG16}, to more specific ones aimed at mitigating racial biases~\cite{DBLP:journals/pami/WangZD22}, all contributed to the development and training of novel approaches. However, the data collection process of such datasets has not followed the best practices with regards to privacy and ethical considerations~\cite{DBLP:conf/icb/BoutrosHSRD22}. Hence, consent is only one of the several concerns regarding FR datasets, which led to the retraction of several of these datasets~\cite{DBLP:journals/corr/abs-2404-04580}.  

FR with synthetic data is a relatively new topic that branched and is now progressing with a lot of attention~\cite{deandres2024frcsyn}. This focuses on fighting two of the main researchers problems. 1) Collecting new data is expensive, and synthetic data can support an artificial increase in the number of samples; 2) With the public retraction of real datasets, the availability of training data reduced drastically, so synthetic datasets could provide a more ethical strategy to train FR systems~\cite{DBLP:journals/ivc/BoutrosSFD23}. Being at its youth, synthetic datasets have already solved several problems, such as having more than one image per identity~\cite{DBLP:conf/iccv/BoutrosGKD23} (ensuring intra-class variability), and generating high-quality samples~\cite{DBLP:conf/cvpr/Kim00023}. Nonetheless, many more remain unsolved as FR models trained on synthetic datasets have lower performances on test datasets~\cite{DBLP:conf/iccv/BoutrosGKD23}, and, thus, can also be related to an increase in the bias~\cite{deandres2024frcsyn}. Differently from other domains, bias in face verification is related to intra-class hardness. For instance, if pairs of images from one ethnicity are easier to classify (as the same person, or not) in comparison to pairs of other group, then the model is biased against the latter.  

Frequently used as a compression strategy, Knowledge Distillation (KD), figures as one potential alternative to support the development of FR systems on synthetic, or partially synthetic datasets~\cite{DBLP:journals/corr/abs-2401-10139}. Given that a dataset has been retracted from the public domain, one cannot use it to train a model. However, if a current model already exists, one can use that model to distil its knowledge to a new model, which can follow the exact same architecture or be slightly different~\cite{boutros2024adadistill}. This ensures that new models do not see real samples, but at the same time, they can leverage the knowledge extracted by them. To the model previously trained on real data, which is usually more complex, we refer to as Teacher, whereas the new model with smaller capacity is the Student.  

Using KD to distil knowledge to a student architecture is a straightforward process. Yet, there are some considerations that should be taken into account. 1) Is it possible to fully replace real data with synthetic one by using KD? 2) What is the impact on the bias of the student model? 3) How much does the performance differ between models trained from scratch in comparison to distilled models? We aim to answer all these questions within this study, exploring these points on different losses~\cite{DBLP:conf/cvpr/BoutrosDKK22,DBLP:conf/cvpr/Kim0L22}, architectures~\cite{DBLP:conf/cvpr/HeZRS16}, train datasets~\cite{DBLP:conf/iccv/BoutrosGKD23,DBLP:conf/cvpr/Kim00023,DBLP:journals/pami/WangZD22} and test sets~\cite{DBLP:conf/iccv/WangDHTH19,LFWTech}. 

Following the above-mentioned elements, we present the following as the main contributions of this research work: 

\begin{itemize}
    \item A mixture of several synthetic datasets based on ethnic balance.
    \item Investigation of training different architectures with different losses on a real datasets, a synthetic dataset and one custom dataset that mixes real and synthetic samples.
    \item A study on the advantages of using KD to distil information through a training with mix and synthetic data.
    \item A validation of the fairness of these approaches and their individual contributions to a bias reduction.
    \item The identification of models trained with fully synthetic data as most affected by KD strategies.
\end{itemize}

We structured this paper in four main sections and two introductory and conclusive sections. The four sections are Related Work, Methodology, Experimental Setup and Results. They are dedicated to describing previous works related to this study and their limitations, describing the proposed methodology, following the experimental setup that allows us to reproduce our results, and finally, highlighting the main findings. Additionally, we are releasing the code and the respective indications of which samples belong to each dataset used. 

\section{Related work}
\subsection{Synthetic Data in Face Recognition}
Due to privacy concerns, recent work has considered using synthetic data in face recognition problems, which has been emphasised by the retraction of large face datasets. However, the challenge of training the models with synthetic data lies in lower performance in comparison to the models trained on real data. Efforts have been made to mitigate such behaviour, with researchers focusing on improving the variability and the quality of data generated. Initial experiments leveraged the use of generative adversarial networks (GAN)~\cite{DBLP:conf/icb/BoutrosHSRD22,DBLP:conf/icb/BoutrosKFKD23}, but these were notoriously hard to train and generated lower quality samples. This has been followed by recent efforts with diffusion models, such as DCFace~\cite{DBLP:conf/cvpr/Kim00023} and IDiff-Face~\cite{DBLP:conf/iccv/BoutrosGKD23}, leading to higher quality results. These efforts led to the creation of two competitions, one targeted at improving fairness with synthetic samples while reducing the gap to models train on real data~\cite{deandres2024frcsyn}, whereas the other focused on the generation of new synthetic data~\cite{DBLP:journals/corr/abs-2404-04580}. Microsoft proposed a graphically generated dataset of faces to train FR systems~\cite{DBLP:conf/wacv/BaeGBHCVCS23}. Atzori~\textit{et al.}~\cite{DBLP:journals/corr/abs-2404-03537} further studied the impact of replacing real data with synthetic samples. Although synthetic data can be a good starting point for reducing different ethical concerns, performance of models trained on them lags behind the one trained on real data. Therefore, combining them with existing techniques for performance improvement can even further highlight their potential.

\subsection{Knowledge Distillation in Face Recognition}
Knowledge distillation has been used as powerful technique to train a model while aiming at either quality improvement, model compression or training with limited data. KD aiming at model compression can be especially useful in biometrics, where models should work on smaller devices~\cite{DBLP:journals/corr/abs-2401-10139, DBLP:journals/access/BoutrosSKDKK22}. This might be accompanied by performance degradation in comparison with the larger model, but improved performance in comparison to the smaller model trained from scratch~\cite{DBLP:journals/corr/abs-2401-10139}. Furthermore, some works have even introduced different tasks or novel capabilities to the student model~\cite{DBLP:journals/corr/abs-2404-09555}.

The work done on face recognition and KD, has shown relevance in the improvement of these models. AdaDistill~\cite{boutros2024adadistill} and SynthDistill~\cite{DBLP:conf/icb/OtroshiShahrezaGM23} have shown significant progress with regards to performance improvements on KD for Face Recognition. Nonetheless, they did not cover the fairness aspects of their approach. Huang~\textit{et al.}~\cite{DBLP:conf/cvpr/HuangWXD22} work was not tested on bias mitigation, but stands as one of the most promising strategies for KD through the optimization on the evaluation space.
Dhar~\textit{et al.}~\cite{DBLP:journals/corr/abs-2112-09786} convert demographic bias within the contex of KD for FR, however, their strategy might be affected by different teacher/student architectures, and does not leverage synthetic data. Jung~\textit{et al.}~\cite{DBLP:conf/cvpr/JungLPM21} designed a strategy for bias mitigation through KD and synthetic data, but did not target FR. Neto~\textit{et al.}~\cite{DBLP:conf/biosig/NetoCCS23} studied the behaviour of bias under other compression techniques such as quantization. Liu~\textit{et al.}~\cite{DBLP:conf/iccvw/LiuZSYL21} explores a strategy based on the uniformity of the training data. However, their approach is not specifically designed for demographic bias, nor does it include the use of synthetic data. As seen, the integration of KD as a strategy for bias mitigation is relatively unexplored within the field of FR. This becomes even more relevant when it is narrowed to FR based on synthetic data.  

To the extent of our knowledge, no other work explored the combination of synthetic data, ethnic-aware sampling and knowledge distillation to create models that have increased fairness and privacy without significant compromises on the performance. Hence, we illustrate the benefits of these strategies through an extensive set of experiments on different architectures. In addition, we design all the experiments to consider model compression as it is one of the benefits linked to KD. 


\section{Methodology}

\begin{figure}
    \centering
    \includegraphics[width=0.95\linewidth]{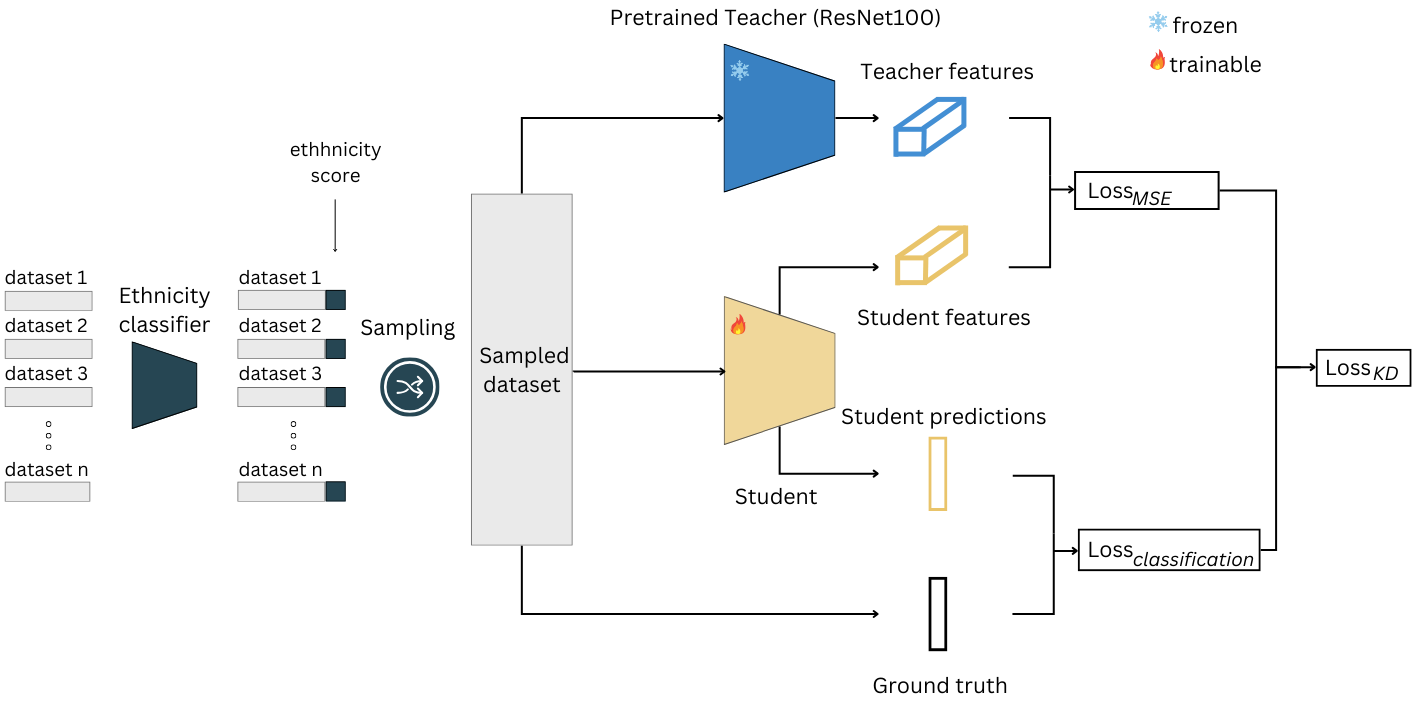}
    \caption{Visual representation of the proposed approach. Includes the merge of several real or synthetic datasets using a ethnicity aware sampling approach. On the top, it is visible a frozen (snowflake icon) pretrained teacher that distils knowledge to a training student (fire icon). }
    \label{fig:methodology}
\end{figure}

Acknowledging the diverse advantages of using knowledge distillation to train new models, we have designed a set of methodological strategies to assess empirically these advantages. Our strategies focus on several aspects such as: various degrees of model compression, the influence of having a dataset composed of real data, synthetic data or a mix of both, and the fairness impact of each of the previous elements. Fundamentally, the following subsections and subsequent experiments were designed to answer the following questions: "Does KD significantly improves models trained with Synthetic Data?'' and "In a scenario where only parts of real data is missing, is a mixture of synthetic and real data beneficial?''. Figure~\ref{fig:methodology} shows a visualisation of the proposed strategy described in this work.

\subsection{Teacher}

Knowledge distillation, also known as student-teacher architecture, is a technique for transferring knowledge from one model to another, even when they do not have the same architecture. It has been shown it can be useful for different objectives, improving performance, model compression or learning with unlabeled data. Due to the resource-constrained environments in which FR systems run, KD is usually used for model compression. Knowledge is distilled from the large neural network called teacher, to smaller neural network called student.

The teacher follows the the standard for recent Face Recognition methods~\cite{DBLP:conf/cvpr/BoutrosDKK22} and is a ResNet100. This teacher is trained with a classification layer that is removed from the final model, resulting in a backbone that produces an embedding vector and no classification output. Hence, it is not possible to use response-based KD strategies that focus on the approximation of soft-probabilities.

\subsection{Student}

Student model is neural network to which knowledge is being distilled. It can have the same or smaller capacity, while it can also have the same or different architecture from teacher. As the aim of this study is model compression, student model is smaller neural network that can be deployed to smaller devices. While we use two progressively smaller architectures known as ResNet-34~\cite{DBLP:conf/cvpr/HeZRS16} and MobileFaceNet~\cite{DBLP:conf/ccbr/ChenLGH18} as student models, other architectures can also be used. These were adapted to have the same embedding dimension as the teacher, so that feature-based KD can be performed.

As KD is being performed at the feature level, the presence of a classification layer is negligible. This allows for KD training with different datasets than the one utilised to train the teacher. This is a major advantage, especially after considering the backlash of face recognition datasets.

In our KD strategy we minimise a function of the KD error and the classification loss. The classification loss chosen is the exact same classification used in the teacher, so if the teacher is trained with AdaFace, the same loss will be used on the student. The knowledge distillation loss, $L_{KD}$, is calculated as the minimum squared error between the embedding produced by the teacher, $\phi_{t}(f_{t})$, and the one produced by the student, $\phi_{t}(f_{s})$. This latter term is weighted with a parameter $\lambda$. Equation \ref{eq:kd_loss} shows the function that gives the final loss value, where the classification loss ($L_{classification}$), either AdaFace or ElasticArcFace, receives $\phi_{s}(f_{s})$ and computes the respective loss, which is related to the angle between this embedding and the weights connecting to the correct output, $y$. 

\begin{equation}
    \label{eq:kd_loss}
    \centering
    loss = L_{classification}(\phi_{s}(f_{s}), y) + \lambda L_{KD}(\phi_{t}(f_{t}), \phi_{s}(f_{s}))
\end{equation}

\subsection{Sampling}

Combining several datasets requires an understanding of the complex nature and distribution of each dataset. In practice, it is not possible to account for all of these elements while computing this step in with an efficient strategy. Hence, we have decided to focus our sampling strategy to the merge of multiple synthetic, or real datasets based on ethnicity information. 

In order to combine these datasets while promoting a balance between ethnicity groups we decided to concatenate them together and follow the sampling strategy proposed by Neto~\textit{et al.}~\cite{neto2024beyond}. For this, we have utilise the ethnicity classifier trained by Neto~\textit{et al.} to compute soft ethnicity labels for each image in the concatenated dataset. This is then used to compute a score for all identities and this score is leveraged to choose which identities are removed. Focusing on this, the final dataset retains the samples that are the most representative of a given ethnicity, leading to samples that contribute more to the learning process. 

The ability to seemly combine several synthetic datasets supports the growth of the field by exploring the strengths of each of the datasets, increasing the variability of the resulting dataset and mitigating the flaws of other datasets. This is the case as will be showcased by the results section, where the merge of two synthetic datasets retains the performance of the best performing one while reducing the bias. 

As training models only on synthetic data is currently not comparable with models trained on real data, we investigate how the translation to synthetic data can be done progressively by utilising a mixture of synthetic and real data. For the purpose of creating a mix of data, several synthetic datasets and a real dataset were concatenated, and Neto~\textit{et al.}~\cite{neto2024beyond} sampling approach was used to reduce the number of identities while retaining ethnicity-aware balance.  




\section{Experimental Setup}

\subsection{Datasets}
\subsubsection{Fair face recognition datasets}

Wang~\textit{et al.}~\cite{DBLP:conf/iccv/WangDHTH19} introduced Racial Faces in-the-Wild (RFW) in 2019, an evaluation dataset composed of four different ethnicity groups - African, Caucasian, Asian and Indian. Their goal, was to create one protocol per ethnicity which led to the construction of 6K pairs of images of the same group. This leads to four sets of 6K pairs with equal number of positive and negative pairs within each set. A model can, due to this dataset, be evaluated independently for each set and the results can be compared to detect potential biased behaviour. In total, it contains 24K different pairs.

Later in 2022, Wand~\textit{et al.}~\cite{DBLP:journals/pami/WangZD22} proposed BUPT-BalancedFace, a training dataset composed of 28K different identities and roughly 1.3M images from these identities. This dataset has the particularity of being balanced with respect to the ethnicity label, so each of the four considered ethnicity groups has images from 7K different identities. This dataset will be used to train the teacher and some instances of the student, and will be referred to as the \textbf{Real Dataset}.

\subsubsection{Synthetic Datasets}

To train the various student models, we have decided to follow with two synthetic datasets that have been generated through a diffusion process. First, we start by using IDiff-Face~\cite{DBLP:conf/iccv/BoutrosGKD23}, which contains two sets of 10K identities with 50 images each. We decided to join both sets totalling 20K identities and 1M images. In addition we also use DCFace~\cite{DBLP:conf/cvpr/Kim00023}, containing 60K identities and 1.3M images. Leveraging Neto~\textit{et al.}~\cite{neto2024beyond} ethnicity-aware sampling protocol, we propose a synthetic dataset that merges identities from both sets totalling 27K identities and 1.2M images. We also show that this achieves comparable performance to DCFace with 2x less identities and far less bias. We will refer to this as \textbf{Synthetic Dataset}.

\begin{figure}[hbtp]
\centering
    \begin{subfigure}{0.2\textwidth}
      \centering
      \includegraphics[width=.7\linewidth]{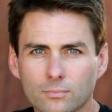}
      \caption{1a}
      \refstepcounter{subfigure}\label{fig:sfig1}
    \end{subfigure}
    \begin{subfigure}{0.2\textwidth}
      \centering
      \includegraphics[width=.7\linewidth]{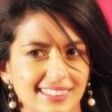}
      \caption{1b}
      \refstepcounter{subfigure}\label{fig:sfig2}
    \end{subfigure}
    \begin{subfigure}{0.2\textwidth}
      \centering
      \includegraphics[width=.7\linewidth]{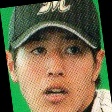}
      \caption{1c}
      \refstepcounter{subfigure}\label{fig:sfig3}
    \end{subfigure}
    \begin{subfigure}{0.2\textwidth}
      \centering
      \includegraphics[width=.7\linewidth]{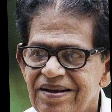}
      \caption{1d}
      \refstepcounter{subfigure}\label{fig:sfig4}
    \end{subfigure}

    \begin{subfigure}{0.2\textwidth}
      \centering
      \includegraphics[width=.7\linewidth]{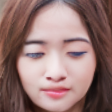}
      \caption{2a}
      \refstepcounter{subfigure}\label{fig:sfig5}
    \end{subfigure}
    \begin{subfigure}{0.2\textwidth}
      \centering
      \includegraphics[width=.7\linewidth]{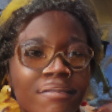}
      \caption{2b}
      \refstepcounter{subfigure}\label{fig:sfig6}
    \end{subfigure}
    \begin{subfigure}{0.2\textwidth}
      \centering
      \includegraphics[width=.7\linewidth]{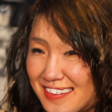}
      \caption{2c}
      \refstepcounter{subfigure}\label{fig:sfig7}
    \end{subfigure}
    \begin{subfigure}{0.2\textwidth}
      \centering
      \includegraphics[width=.7\linewidth]{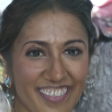}
      \caption{2d}
      \refstepcounter{subfigure}\label{fig:sfig8}
    \end{subfigure}

    \begin{subfigure}{0.2\textwidth}
      \centering
      \includegraphics[width=.7\linewidth]{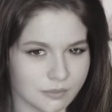}
      \caption{3a}
      \refstepcounter{subfigure}\label{fig:sfig9}
    \end{subfigure}
    \begin{subfigure}{0.2\textwidth}
      \centering
      \includegraphics[width=.7\linewidth]{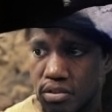}
      \caption{3b}
      \refstepcounter{subfigure}\label{fig:sfig10}
    \end{subfigure}
    \begin{subfigure}{0.2\textwidth}
      \centering
      \includegraphics[width=.7\linewidth]{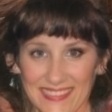}
      \caption{3c}
      \refstepcounter{subfigure}\label{fig:sfig11}
    \end{subfigure}
    \begin{subfigure}{0.2\textwidth}
      \centering
      \includegraphics[width=.7\linewidth]{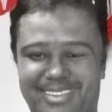}
      \caption{3d}
      \refstepcounter{subfigure}\label{fig:sfig12}
    \end{subfigure}
    \caption{Samples of images that appear in all datasets retrieved from BalancedFace (real data) in the first row, IDiff-Face (synthetic) in the second row and DCFace (synthetic) in the third row.}
    \label{fig:fig_samples}
\end{figure}

We further introduce a third training dataset, referred to as \textbf{Mix Dataset}, which is a combination of the three datasets, DCFace, IDiff-Face and BUPT-Balanced. This should represent a middle step between where instead of removing all real data we gradually replace it with synthetic samples. This set contains 1.2M images from 22K identities. \~70\% of images/IDs are real, while \~30\% are synthetic. Example of images from BUPT-BalancedFace, IDiffFace and DCFace used in all three datasets (Real, Synthetic, Mix) can be seen in Figure \ref{fig:fig_samples}   

\subsubsection{Face recognition datasets}
LFW~\cite{LFWTech} dataset contains 13K images from more than 5K identities organised in 6K pairs with equal number of positive and negative pairs. Cross-pose LFW (CPLFW)~\cite{zheng2018cross} contains 11K images of 4K identities containing various poses. These have been combined in pairs and have been used to assess the performance under a more difficult scenario of cross-poses. Similarly, Cross-Age LFW (CALFW)~\cite{DBLP:journals/corr/abs-1708-08197} dataset considers different age distributions to evaluate the performance of the model when exposed to pairs with wide age gaps. Additionally, we have also tested on a second age-aware dataset called AgeDB~\cite{DBLP:conf/cvpr/MoschoglouPSDKZ17}. Further validation of the cross-pose performance using CFP-FP~\cite{DBLP:conf/wacv/SenguptaCCPCJ16} dataset was done, which contains images of celebrities in frontal and profile views.

\subsection{Experimental Details}

We have trained four different architectures, ResNet-100, -50, -34~\cite{DBLP:conf/cvpr/HeZRS16} and MobileFaceNet~\cite{DBLP:conf/ccbr/ChenLGH18}. Each has been trained with both AdaFace~\cite{DBLP:conf/cvpr/Kim0L22} and ElasticArcFace~\cite{DBLP:conf/cvpr/BoutrosDKK22}. AdaFace parameters have been set to $s=60$ and $m=0.4$, while ElasticArcFace parameters were set to $s=64$, $m=0.5$ and $std=0.05$. All architectures output an embedding with 512 dimensions, which was chosen to be similar across architectures (for distillation). Despite Boutros~\textit{et al.}~\cite{DBLP:conf/iccv/BoutrosGKD23} suggestion to use RandAugmentations when synthetic dataset are used, we have decided to leverage only random horizontal flips to stay consistent with the training of the teacher. Bare in mind that such augmentations could boost the results of models distilled with synthetic data even further. Each model was trained for 26 epochs, with a batch size of 256 and weights were optimised with stochastic gradient descent (SGD). The initial learning rate of 0.1 decayed by a factor of 10 at epochs 8, 14, 20 and 25. All models have been trained on a server with one NVIDIA A100 80GB GPU. The sampling strategy was set so that all datasets had a similar number of images. Besides ResNet-100, which is used as the teacher and is only trained from scratch on real data, all the other models have been trained from scratch and with KD for all training datasets. The code is publicly available on GitHub~\footnote{https://github.com/ivonacolakovic/SynthGap-mitigation-using-KD-in-FFR}.

\subsection{Evaluation Metrics}

Due to the division of RFW into different ethnicity groups, we can individually calculate the accuracy for each of these groups. Hence, we can measure their accuracy individually and globally through the average of the different accuracies. In addition, inter-ethnicity bias can be measured by calculating additional metrics. For instance, the standard deviation of sensitive feature groups' accuracy (STD) and skewed error ratio (SER)~\cite{DBLP:journals/pami/WangZD22} give us insights regarding the fairness of the model with respect to these demographic groups. SER represents the ratio of the highest error rate to the lowest error rate and is calculated as shown in Equation \ref{eq:ser}, where $g$ represents ethnicity group.

\begin{equation}
    \label{eq:ser}
    \centering
    SER = \frac{100-\underset{g}{min}(acc_{g})} {100-\underset{g}{max}(acc_{g})}
\end{equation}

Both STD and SER indicate increased fairness as their values decrease, meaning that the best value of STD and SER is 0, while the worst is 1. For the remaining evaluation datasets, only accuracy is measured due to the lack of ethnicity labels. 

\section{Results}

\subsection{Ablation on Dataset Merge}

We start by assessing the effectiveness of the sampling strategy on the merge of synthetic datasets. We show, in Table~\ref{tab:results_merge} the performance of a ResNet-34 and MobileFaceNet trained on the concatenation of both IDiff-Face datasets, DCFace, a version of DCFace which is sampled down to 20K identities and our Synthetic dataset that merges IDiff-Face and DCFace.

\begin{table}[h!]
\caption{Comparison between different datasets used to train the same architecture with AdaFace. Synthetic represents the merge between IDiff-Face and DCFace, where DCFace (Sampled) represents a sampling exclusively done with DCFace data. Results on RFW. Best results in bold. }
\label{tab:results_merge}
\centering
\resizebox{\textwidth}{!}{%
\begin{tabular}{@{}lccccccccc@{}}
\toprule
 &  &  & \multicolumn{5}{c}{Accuracy (\%)} & \multicolumn{1}{l}{} & \multicolumn{1}{l}{} \\ \cmidrule(lr){4-8}
Model & \# IDs & Data & \multicolumn{1}{l}{Caucasian} & \multicolumn{1}{l}{Indian} & \multicolumn{1}{l}{Asian} & \multicolumn{1}{l}{African} & \multicolumn{1}{l}{Average} & \multicolumn{1}{l}{STD} & \multicolumn{1}{l}{SER} \\ \midrule
ResNet 34 & 27K & Synthetic &  88.87 & 86.02 & \textbf{83.07} & \textbf{80.53} & 84.62 & \textbf{3.61} & 1.75  \\
ResNet 34 & 60K & DCFace & \textbf{89.43} & \textbf{86.27} & 82.88 & 79.95 & \textbf{84.63} & 4.11 & 1.90 \\
ResNet 34 & 20K & DCFace (Sampled) & 88.27 & 85.02 & 81.97 & 79.22 & 83.62 & 3.90 & 1.77 \\
ResNet 34 & 20K & IDiff-Face & 84.87 & 80.77 & 79.20 & 75.63 & 80.12 & 3.83 & \textbf{1.61} \\ \midrule
MobileFaceNet & 27K & Synthetic & 86.67 & 83.75 & 80.15 & 77.35 & 81.98 & \textbf{4.08} & 1.70 \\
MobileFaceNet & 60K & DCFace & \textbf{87.8} & \textbf{84.32} & \textbf{80.95} & \textbf{77.98} & \textbf{82.76} & 4.24 & 1.80 \\
MobileFaceNet & 20K & DCFace (Sampled) & 86.63 & 83.38 & 80.3 & 77.08 & 81.85 & 4.10 & 1.71 \\
MobileFaceNet & 20K & IDiff-Face & 83.20 & 79.70 & 77.47 & 72.75 & 78.28 & 4.38 & \textbf{1.62} \\ \bottomrule
\end{tabular}%
}
\end{table}

Our results show that IDiff-Face performs poorly in comparison to the other approaches on this evaluation dataset. DCFace showcases a strong performance, but the displayed very high bias values. The sampled version of DCFace shows slight improvements on bias metrics, but at the cost of 1 percentual point on average accuracy. On the other hand, the merged dataset (Synthetic), with roughly 2x less identities, achieves the same performance of DCFace while reducing the STD by 13\%. Although on MobileFaceNet best results are achieved with DCFace, it must be taken into account that DCFace has three times more samples than the rest and it still performs poorly regarding fairness.

These results highlight the advantages of merging a two datasets, even if one (IDiff-Face) achieves significantly lower performance.

\subsection{Validation on Low-Compression}
Using KD as a compression tool can have some challenges with possible complexity gap between teacher and student among others. By using high-capacity teacher to distil knowledge to much smaller student, fine-grained data can be lost. To investigate how the capacity of the student network impacts both overall and ethnicity-based performance, we perform additional experiments using ResNet 50 as a medium-sized neural network and AdaFace loss. 

The results shown in Table \ref{tab:results_resnet50}, reveal that KD using mix dataset achieves close results to the model trained from scratch on the real dataset, while still having a small gap when comparing to KD using the real dataset. By being only \~3\% behind other two approaches, KD with only synthetic data on medium size network is much closer to others' performance compared to when small student models is used. The fairness of model is also improved compared to smaller models. Nevertheless, by observing KD with mix data the improvement of both fairness metrics can be noticed when compared to the KD with real data. Interestingly, fairness is also better when compared to the same model (ResNet 50) trained from scratch on real data. Therefore, it can be concluded that trade-off between computational complexity and fairness should be considered when utilising KD in FR, while additional attention should be paid to fairness when student has small capacity.

\begin{table}[hbtp]
\caption{Accuracy per ethnicity and overall results of KD performed with sampled datasets using ResNet-50 and AdaFace loss on RFW dataset.}
\label{tab:results_resnet50}
\centering
\resizebox{\textwidth}{!}{%
\begin{tabular}{@{}lccccccccc@{}}
\toprule
 &  &  & \multicolumn{5}{c}{Accuracy (\%)} & \multicolumn{1}{l}{} & \multicolumn{1}{l}{} \\ \cmidrule(lr){4-8}
Model & Distillation & Data & \multicolumn{1}{l}{Caucasian} & \multicolumn{1}{l}{Indian} & \multicolumn{1}{l}{Asian} & \multicolumn{1}{l}{African} & \multicolumn{1}{l}{Average} & \multicolumn{1}{l}{STD} & \multicolumn{1}{l}{SER} \\ \midrule
ResNet 100 & No & Real & 97.40 & 96.07 & 95.52 & 95.95 & 96.24 & 0.81 & 1.72 \\
ResNet 50 & No & Real & 97.30 & 95.78 & 95.15 & 95.13 & 95.84 & 1.02 & 1.80 \\
ResNet 50 & Yes & Real & 97.73 & 96.27 & 95.57 & 95.93 & 96.38 & 0.95 & 1.95 \\
ResNet 50 & Yes & Synthetic & 95.63 & 93.20 & 92.25 & 91.55 & 93.16 & 1.78 & 1.93 \\
ResNet 50 & Yes & Mix & 97.05 & 96.12 & 95.27 & 95.08 & 95.88 & 0.90 & 1.67 \\ \bottomrule
\end{tabular}%
}
\end{table}

\subsection{Fairness Evaluation}

Results showing how knowledge distillation with different data impacts the verification performance compared to the model trained from scratch are presented in Table \ref{tab:results_fair}. The table shows accuracy per ethnicity, overall accuracy, and two fairness metrics STD and SER. With the superior performance of ResNet100 in all regards, it is suitable for it to be a teacher model. Although the distillation to the ResNet34 with both losses achieves worse results when only synthetic data is used, results with a mix of synthetic and real data are comparable to the distillation with real data in both accuracy and fairness. The same can be observed when MobileFaceNet is used. ResNet34 with distillation done with mix data and ElasticArcFace loss even improves SER (1.73 compared to 1.85), as well as MobileFaceNet with AdaFace (1.60 compared to 1.65). While using AdaFace as loss generally improves results when compared to ElasticArcFace, it does not impact the results disproportionately.

\begin{table}[h!]
\caption{Accuracy per ethnicity and overall results of the teacher, the baseline and proposed methods on RFW dataset. Results on different architectures, datasets, losses and training strategies. }
\label{tab:results_fair}
\centering
\resizebox{\textwidth}{!}{%
\begin{tabular}{@{}llccccccccc@{}}
\toprule
 &  &  &  & \multicolumn{5}{c}{Accuracy (\%)} & \multicolumn{1}{l}{} & \multicolumn{1}{l}{} \\ \cmidrule(lr){5-9}
Loss & Model & Distillation & Data & \multicolumn{1}{l}{Caucasian} & \multicolumn{1}{l}{Indian} & \multicolumn{1}{l}{Asian} & \multicolumn{1}{l}{African} & \multicolumn{1}{l}{Average} & \multicolumn{1}{l}{STD} & \multicolumn{1}{l}{SER} \\ \midrule
 & ResNet 100 & No & Real & 97.12 & 95.78 & 94.93 & 95.36 & 95.80 & 0.95 & 1.76 \\
 & ResNet 34 & No & Real & 96.67 & 94.88 & 94.22 & 93.38 & 94.79 & 1.40 & 1.99 \\
 & ResNet 34 & No & Synthetic & 88.78 & 84.82 & 81.83 & 79.43 & 83.72 & 4.03 & 1.83 \\
 & ResNet 34 & No & Mix & 94.75 & 93.82 & 92.00 & 90.48 & 92.76 & 1.90 & 1.81 \\ \cmidrule(l){4-11} 
 & ResNet 34 & Yes & Real & 96.98 & 95.12 & 94.73 & 94.42 & 95.31 & 1.15 & 1.85 \\
 & ResNet 34 & Yes & Synthetic & 94.02 & 90.82 & 90.33 & 88.18 & 90.84 & 2.41 & 1.98 \\
ElasticArcFace & ResNet 34 & Yes & Mix & 96.12 & 94.87 & 93.87 & 93.27 & 94.53 & 1.25 & 1.73 \\ \cmidrule(l){3-11} 
 & MobileFaceNet & No & Real & 93.37 & 91.82 & 90.28 & 88.98 & 91.11 & 1.90 & 1.66 \\
 & MobileFaceNet & No & Synthetic & 86.10 & 82.58 & 79.92 & 77.15 & 81.44 & 3.82 & 1.64 \\
 & MobileFaceNet & No & Mix & 92.27 & 91.28 & 88.80 & 86.32 & 89.67 & 2.67 & 1.77 \\ \cmidrule(l){4-11} 
 & MobileFaceNet & Yes & Real & 94.18 & 92.37 & 91.27 & 90.17 & 92.00 & 1.71 & 1.69 \\
 & MobileFaceNet & Yes & Synthetic & 90.95 & 87.92 & 85.75 & 83.00 & 86.91 & 3.37 & 1.88 \\
 & MobileFaceNet & Yes & Mix & 93.68 & 91.67 & 90.55 & 88.67 & 91.14 & 2.10 & 1.79 \\ \cmidrule(l){1-11} 
 & ResNet 100 & No & Real & 97.40 & 96.07 & 95.52 & 95.95 & 96.24 & 0.81 & 1.72 \\
 & ResNet 34 & No & Real & 96.73 & 95.30 & 94.68 & 94.00 & 95.18 & 1.16 & 1.83 \\
 & ResNet 34 & No & Synthetic & 88.87 & 86.02 & 83.07 & 80.53 & 84.62 & 3.61 & 1.75 \\
 & ResNet 34 & No & Mix & 94.88 & 94.03 & 92.62 & 90.98 & 93.13 & 1.71 & 1.76 \\ \cmidrule(l){4-11} 
 & ResNet 34 & Yes & Real & 96.92 & 95.88 & 94.98 & 95.20 & 95.75 & 0.87 & 1.63 \\
 & ResNet 34 & Yes & Synthetic & 95.28 & 92.67 & 91.55 & 90.05 & 92.39 & 2.21 & 2.11 \\
AdaFace & ResNet 34 & Yes & Mix & 96.78 & 95.18 & 94.50 & 94.37 & 95.21 & 1.11 & 1.75 \\ \cmidrule(l){3-11} 
 & MobileFaceNet & No & Real & 93.48 & 92.13 & 90.12 & 88.80 & 91.13 & 2.08 & 1.72 \\
 & MobileFaceNet & No & Synthetic & 86.67 & 83.75 & 80.15 & 77.35 & 81.98 & 4.08 & 1.70 \\
 & MobileFaceNet & No & Mix & 92.23 & 91.25 & 88.68 & 87.00 & 89.79 & 2.39 & 1.67 \\ \cmidrule(l){4-11} 
 & MobileFaceNet & Yes & Real & 94.15 & 92.15 & 91.65 & 90.37 & 92.08 & 1.57 & 1.65 \\
 & MobileFaceNet & Yes & Synthetic & 91.80 & 88.85 & 87.03 & 84.92 & 88.15 & 2.92 & 1.84 \\
 & MobileFaceNet & Yes & Mix & 93.65 & 92.28 & 90.88 & 89.87 & 91.67 & 1.65 & 1.60 \\ \bottomrule
\end{tabular}%
}
\end{table}

When comparing student models with models trained from scratch, it can be observed that the biggest improvement of distillation is when it is done with only synthetic data, since the increase is around 5 to 8 percent. The notable improvement in results given by using synthetic data in distillation is a promising fact considering that, currently, training models from scratch with synthetic data does not give similar performance. Using KD with real data improves average accuracy for at least 0.5 percent points, while KD with mix data achieves \~2\% better accuracy. Regarding fairness, KD improves both STD and SER when real and mix data are used. On the other hand, while KD can help to improve STD notably, it tends to hurt SER when only synthetic data is used, no matter which model and loss are utilized.

By observing the results achieved by each ethnicity individually, it can be seen that Caucasian samples perform the best and African the worst, while Indian and Asian groups are comparable. While KD with mixed data improves results in a similar manner for all groups when compared to KD with real data, KD with synthetic data gives the biggest boost in African accuracy, where they usually have 2\% higher accuracy than other groups. Interestingly, performing KD with mix data using MobileFaceNet with AdaFace loss, Indian achieved higher accuracy than KD with real data. The same model, shows the biggest improvement in accuracy of Asian samples, although they are comparable to the other groups.

The comparison of student models to the same architecture networks trained from scratch, we noticed a comparable improvement for all ethnicities when the student is learning by distilling the knowledge with real data. A small difference is present in KD with mix data, where African accuracy is higher for at least 2\%, while other groups accuracy usually improves for less than 2\%. The biggest difference in accuracy between groups after the distillation is evident when only synthetic data is used. The most noticeable improvement is visible in Asian and African group when ResNet34 is student model, where the improvement is at least 8.5\% in both groups. Caucasian and Indian groups achieve similar improvement by having \~5\% in MobileFaceNet and \~6\% in ResNet34 accuracy improvement. It is noteworthy how distillation enhances accuracy more significantly for African and Asian groups when only synthetic data is utilised, whereas it improves accuracy for African groups more than for other groups when mixed data is utilised.

\subsection{Performance Evaluation}

To further evaluate and analyse the proposed approach, the evaluation is done on multiple face recognition datasets, and results are shown in Table \ref{tab:results}. While the biggest differences in performance can be seen on RFW and CFP\_FP dataset, results on LFW dataset show that the best accuracy of student model is achieved when KD is done with mix data and KD with synthetic data boosts accuracy so it is comparable to the other  approaches. Accuracy of students' models on other datasets achieve comparable results, where, as seen before, ResNet34 achieves higher accuracy than MobileFaceNet.

It is interesting how ResNet34 using distillation with only synthetic data achieves higher accuracy than the MobileFaceNet using distillation with real data in 4 out of 6 datasets. This means that with a slight increase in computational complexity by using ResNet34 instead of MobileFaceNet, privacy can be preserved without losing quality.

\begin{table}[hbtp]
\caption{Accuracy results of the teacher, the baseline and proposed methods with AdaFace loss on different face recognition datasets. Results on different architectures, datasets and training strategies.}
\label{tab:results}
\centering
\resizebox{\textwidth}{!}{%
\begin{tabular}{@{}lccccccccr@{}}
\toprule
Model & Distillation & Data & \multicolumn{1}{l}{RFW} & \multicolumn{1}{l}{LFW} & \multicolumn{1}{l}{CPLFW} & \multicolumn{1}{l}{CALFW} & \multicolumn{1}{l}{AgeDB} & \multicolumn{1}{l}{CFP\_FP} & \multicolumn{1}{l}{Avg} \\ \midrule
ResNet 100 & No & Real & 96.24 & 99.72 & 92.65 & 95.77 & 97.23 & 97.81 & 96.64 \\ \cmidrule(l){2-10}
ResNet 34 & No & Real & 95.18 & 99.67 & 91.28 & 95.55 & 96.53 & 96.86 & 95.98 \\
ResNet 34 & No & Synthetic & 84.62 & 98.63 & 83.08 & 92.53 & 90.68 & 84.94 & 89.97 \\
ResNet 34 & No & Mix & 93.13 & 99.60 & 90.23 & 94.98 & 95.82 & 95.70 & 95.27 \\ \cmidrule(l){4-10} 
ResNet 34 & Yes & Real & 95.75 & 99.63 & 92.02 & 95.72 & 96.85 & 97.27 & 96.30 \\
ResNet 34 & Yes & Synthetic & 92.39 & 99.62 & 88.85 & 95.30 & 96.68 & 93.13 & 94.52 \\
ResNet 34 & Yes & Mix & 95.21 & 99.67 & 91.48 & 95.53 & 96.68 & 96.64 & 96.00 \\ \cmidrule(l){3-10} 
MobileFaceNet & No & Real & 91.13 & 99.37 & 89.35 & 94.40 & 94.57 & 93.73 & 94.28 \\
MobileFaceNet & No & Synthetic & 81.98 & 98.20 & 81.77 & 91.78 & 89.10 & 84.04 & 88.98 \\
MobileFaceNet & No & Mix & 89.79 & 99.37 & 87.67 & 93.95 & 93.35 & 93.17 & 93.50 \\ \cmidrule(l){4-10} 
MobileFaceNet & Yes & Real & 92.08 & 99.42 & 89.73 & 95.00 & 95.63 & 93.89 & 94.73 \\
MobileFaceNet & Yes & Synthetic & 88.15 & 99.32 & 86.10 & 93.93 & 93.63 & 89.63 & 92.52 \\
MobileFaceNet & Yes & Mix & 91.67 & 99.43 & 88.97 & 94.70 & 95.12 & 93.49 & 94.34 \\ \bottomrule
\end{tabular}%
}
\end{table}

\section{Conclusion}

In this study, we explored the potential of knowledge distillation (KD) in bridging the performance gap between synthetic and real data for face recognition tasks. By employing a ResNet100 model, trained on the BalancedFace dataset as the teacher, we investigated the potential of KD to enhance the performance of student models (ResNet34 and MobileFaceNet) using three different types of data: synthetic, real, and a mix of both. The synthetic and mix dataset are obtained by utilising an ethnicity classifier to assign ethnicity score for each sample, based on which it is decided which samples are retrieved for the dataset.

Our findings indicate that while distillation using only synthetic data results in a performance gap when compared to models trained on real data, the introduction of a mixed dataset comprising 70\% real and 30\% synthetic data significantly mitigates this gap. Specifically, we observed that models distilled using the mixed dataset achieve comparable performance in both accuracy and fairness metrics to those distilled using real data alone. This suggests that while training with synthetic data alone does not yield optimal performance, we can progressively move away from real data by using a combination of both synthetic and real data.

KD has been shown to be useful also in fairness terms, as it in some cases even improves SER, highlighting the potential of KD in reducing bias across different ethnic groups. Notably, the use of ElasticArcFace and AdaFace losses contributed to these improvements, with AdaFace generally providing a slight edge over ElasticArcFace.

The ethnic group-specific analysis revealed that KD with synthetic data alone yields the highest accuracy improvement for the African group, which traditionally underperforms compared to other groups. This improvement is crucial in addressing biases inherent in face recognition systems. Furthermore, the mixed dataset approach consistently improved the performance across all ethnic groups, ensuring a more equitable distribution of model accuracy.

Our results on additional face recognition datasets corroborate these findings, demonstrating that student models distilled with mixed data perform comparably to those trained from scratch on real data. Particularly, ResNet34 distilled with synthetic data achieved higher accuracy than MobileFaceNet distilled with real data in four out of six datasets, underscoring the effectiveness of KD in leveraging synthetic data to achieve high performance. This finding poses the question of trade-off between privacy and computational complexity, since we show that by having a little more complex model (ResNet34) compared to MobileFaceNet, we can protect privacy by using synthetic data, while maintaining high performance.

Interestingly, it was shown that by using medium-size student model, performance can be comparable across all datasets, even synthetic, and superior fairness can be achieved with mix dataset also when compared to large-capacity teacher model. This yields new questions for the future work, where use of the medium-size network as a mediator called teaching assistant can be used.

While it was shown that KD can help in mitigating both the performance gap caused by usage of synthetic data and unfairness that appear in FR systems, it has it's limitations. Along with the before-mentioned capacity gap between the teacher and student model, training student models by using the KD technique is more computationally heavy than training them from scratch. However, computational overhead only appears in the training phase, while having only student model in real-time applications allows for fast processing. We find KD techniques useful in improving the small-capacity model's performance despite the increase in computational heaviness in the training phase.

To further solidify the performance of proposed methodology additional experiments with different architectures and loss functions could be conducted. Although it was shown that KD notably improves models trained with only synthetic data, the gap between them and models trained on real data still remains. Mitigating that gap could definitely boost the usage of synthetic data in FR systems, while preserving privacy, thus making them more ethical.

\section*{Acknowledgements}
This work was developed within the Component 5 - Capitalization and Business Innovation, integrated in the Resilience Dimension of the Recovery and Resilience Plan within the scope of the Recovery and Resilience Mechanism (MRR) of the European Union (EU), framed in the Next Generation EU, for the period 2021 - 2026, within project NewSpacePortugal, with reference 11. The authors also acknowledge the financial support from the Slovenian Research and Innovation Agency (Research Core Funding No. P2-0057).


%
%
\bibliographystyle{splncs04}
\bibliography{main}

\begin{thebibliography}{10}
\providecommand{\url}[1]{\texttt{#1}}
\providecommand{\urlprefix}{URL }
\providecommand{\doi}[1]{https://doi.org/#1}

\bibitem{DBLP:journals/corr/abs-2404-03537}
Atzori, A., Boutros, F., Damer, N., Fenu, G., Marras, M.: If it's not enough, make it so: Reducing authentic data demand in face recognition through synthetic faces. CoRR  \textbf{abs/2404.03537} (2024). \doi{10.48550/ARXIV.2404.03537}, \url{https://doi.org/10.48550/arXiv.2404.03537}

\bibitem{DBLP:journals/corr/abs-2404-09555}
Babnik, Z., Boutros, F., Damer, N., Peer, P., Struc, V.: {AI-KD:} towards alignment invariant face image quality assessment using knowledge distillation. CoRR  \textbf{abs/2404.09555} (2024). \doi{10.48550/ARXIV.2404.09555}, \url{https://doi.org/10.48550/arXiv.2404.09555}

\bibitem{DBLP:conf/wacv/BaeGBHCVCS23}
Bae, G., de~La~Gorce, M., Baltrusaitis, T., Hewitt, C., Chen, D., Valentin, J.P.C., Cipolla, R., Shen, J.: Digiface-1m: 1 million digital face images for face recognition. In: {IEEE/CVF} Winter Conference on Applications of Computer Vision, {WACV} 2023, Waikoloa, HI, USA, January 2-7, 2023. pp. 3515--3524. {IEEE} (2023). \doi{10.1109/WACV56688.2023.00352}, \url{https://doi.org/10.1109/WACV56688.2023.00352}

\bibitem{DBLP:conf/cvpr/BoutrosDKK22}
Boutros, F., Damer, N., Kirchbuchner, F., Kuijper, A.: Elasticface: Elastic margin loss for deep face recognition. In: {CVPR} Workshops. pp. 1577--1586. {IEEE} (2022)

\bibitem{DBLP:conf/iccv/BoutrosGKD23}
Boutros, F., Grebe, J.H., Kuijper, A., Damer, N.: Idiff-face: Synthetic-based face recognition through fizzy identity-conditioned diffusion models. In: {ICCV}. pp. 19593--19604. {IEEE} (2023)

\bibitem{DBLP:conf/icb/BoutrosHSRD22}
Boutros, F., Huber, M., Siebke, P., Rieber, T., Damer, N.: Sface: Privacy-friendly and accurate face recognition using synthetic data. In: {IJCB}. pp. 1--11. {IEEE} (2022)

\bibitem{DBLP:conf/icb/BoutrosKFKD23}
Boutros, F., Klemt, M., Fang, M., Kuijper, A., Damer, N.: Exfacegan: Exploring identity directions in gan's learned latent space for synthetic identity generation. In: {IEEE} International Joint Conference on Biometrics, {IJCB} 2023, Ljubljana, Slovenia, September 25-28, 2023. pp. 1--10. {IEEE} (2023). \doi{10.1109/IJCB57857.2023.10449036}, \url{https://doi.org/10.1109/IJCB57857.2023.10449036}

\bibitem{DBLP:journals/access/BoutrosSKDKK22}
Boutros, F., Siebke, P., Klemt, M., Damer, N., Kirchbuchner, F., Kuijper, A.: Pocketnet: Extreme lightweight face recognition network using neural architecture search and multistep knowledge distillation. {IEEE} Access  \textbf{10},  46823--46833 (2022). \doi{10.1109/ACCESS.2022.3170561}, \url{https://doi.org/10.1109/ACCESS.2022.3170561}

\bibitem{boutros2024adadistill}
Boutros, F., {\v{S}}truc, V., Damer, N.: Adadistill: Adaptive knowledge distillation for deep face recognition. ECCV  (2024)

\bibitem{DBLP:journals/ivc/BoutrosSFD23}
Boutros, F., Struc, V., Fi{\'{e}}rrez, J., Damer, N.: Synthetic data for face recognition: Current state and future prospects. Image Vis. Comput.  \textbf{135},  104688 (2023). \doi{10.1016/J.IMAVIS.2023.104688}, \url{https://doi.org/10.1016/j.imavis.2023.104688}

\bibitem{DBLP:journals/corr/abs-2401-10139}
Caldeira, E., Neto, P.C., Huber, M., Damer, N., Sequeira, A.F.: Model compression techniques in biometrics applications: {A} survey. CoRR  \textbf{abs/2401.10139} (2024)

\bibitem{DBLP:conf/ccbr/ChenLGH18}
Chen, S., Liu, Y., Gao, X., Han, Z.: Mobilefacenets: Efficient cnns for accurate real-time face verification on mobile devices. In: {CCBR}. Lecture Notes in Computer Science, vol. 10996, pp. 428--438. Springer (2018)

\bibitem{deandres2024frcsyn}
DeAndres-Tame, I., Tolosana, R., Melzi, P., Vera-Rodriguez, R., Kim, M., Rathgeb, C., Liu, X., Morales, A., Fierrez, J., Ortega-Garcia, J., et~al.: Frcsyn challenge at cvpr 2024: Face recognition challenge in the era of synthetic data. In: Proceedings of the IEEE/CVF Conference on Computer Vision and Pattern Recognition. pp. 3173--3183 (2024)

\bibitem{DBLP:conf/cvpr/DengGXZ19}
Deng, J., Guo, J., Xue, N., Zafeiriou, S.: Arcface: Additive angular margin loss for deep face recognition. In: {CVPR}. pp. 4690--4699. Computer Vision Foundation / {IEEE} (2019)

\bibitem{DBLP:journals/corr/abs-2112-09786}
Dhar, P., Gleason, J., Roy, A., Castillo, C.D., Phillips, P.J., Chellappa, R.: Distill and de-bias: Mitigating bias in face recognition using knowledge distillation. CoRR  \textbf{abs/2112.09786} (2021), \url{https://arxiv.org/abs/2112.09786}

\bibitem{DBLP:conf/eccv/GuoZHHG16}
Guo, Y., Zhang, L., Hu, Y., He, X., Gao, J.: Ms-celeb-1m: {A} dataset and benchmark for large-scale face recognition. In: {ECCV} {(3)}. Lecture Notes in Computer Science, vol.~9907, pp. 87--102. Springer (2016)

\bibitem{DBLP:conf/cvpr/HeZRS16}
He, K., Zhang, X., Ren, S., Sun, J.: Deep residual learning for image recognition. In: {CVPR}. pp. 770--778. {IEEE} Computer Society (2016)

\bibitem{LFWTech}
Huang, G.B., Ramesh, M., Berg, T., Learned-Miller, E.: Labeled faces in the wild: A database for studying face recognition in unconstrained environments. Tech. Rep. 07-49, University of Massachusetts, Amherst (October 2007)

\bibitem{DBLP:conf/cvpr/HuangWXD22}
Huang, Y., Wu, J., Xu, X., Ding, S.: Evaluation-oriented knowledge distillation for deep face recognition. In: {IEEE/CVF} Conference on Computer Vision and Pattern Recognition, {CVPR} 2022, New Orleans, LA, USA, June 18-24, 2022. pp. 18719--18728. {IEEE} (2022). \doi{10.1109/CVPR52688.2022.01818}, \url{https://doi.org/10.1109/CVPR52688.2022.01818}

\bibitem{DBLP:conf/cvpr/JungLPM21}
Jung, S., Lee, D., Park, T., Moon, T.: Fair feature distillation for visual recognition. In: {IEEE} Conference on Computer Vision and Pattern Recognition, {CVPR} 2021, virtual, June 19-25, 2021. pp. 12115--12124. Computer Vision Foundation / {IEEE} (2021). \doi{10.1109/CVPR46437.2021.01194}, \url{https://openaccess.thecvf.com/content/CVPR2021/html/Jung\_Fair\_Feature\_Distillation\_for\_Visual\_Recognition\_CVPR\_2021\_paper.html}

\bibitem{DBLP:conf/cvpr/Kim0L22}
Kim, M., Jain, A.K., Liu, X.: Adaface: Quality adaptive margin for face recognition. In: {CVPR}. pp. 18729--18738. {IEEE} (2022)

\bibitem{DBLP:conf/cvpr/Kim00023}
Kim, M., Liu, F., Jain, A.K., Liu, X.: Dcface: Synthetic face generation with dual condition diffusion model. In: {CVPR}. pp. 12715--12725. {IEEE} (2023)

\bibitem{DBLP:conf/iccvw/LiuZSYL21}
Liu, B., Zhang, S., Song, G., You, H., Liu, Y.: Rectifying the data bias in knowledge distillation. In: {IEEE/CVF} International Conference on Computer Vision Workshops, {ICCVW} 2021, Montreal, BC, Canada, October 11-17, 2021. pp. 1477--1486. {IEEE} (2021). \doi{10.1109/ICCVW54120.2021.00171}, \url{https://doi.org/10.1109/ICCVW54120.2021.00171}

\bibitem{DBLP:conf/cvpr/MoschoglouPSDKZ17}
Moschoglou, S., Papaioannou, A., Sagonas, C., Deng, J., Kotsia, I., Zafeiriou, S.: Agedb: The first manually collected, in-the-wild age database. In: {CVPR} Workshops. pp. 1997--2005. {IEEE} Computer Society (2017)

\bibitem{DBLP:conf/biosig/NetoCCS23}
Neto, P.C., Caldeira, E., Cardoso, J.S., Sequeira, A.F.: Compressed models decompress race biases: What quantized models forget for fair face recognition. In: Damer, N., Gomez{-}Barrero, M., Raja, K.B., Rathgeb, C., Sequeira, A.F., Todisco, M., Uhl, A. (eds.) International Conference of the Biometrics Special Interest Group, {BIOSIG} 2023, Darmstadt, Germany, September 20-22, 2023. pp.~1--5. {IEEE} (2023). \doi{10.1109/BIOSIG58226.2023.10346003}, \url{https://doi.org/10.1109/BIOSIG58226.2023.10346003}

\bibitem{neto2024beyond}
Neto, P.C., Damer, N., Cardoso, J.S., Sequeira, A.F.: Beyond black and white: A more nuanced approach to facial recognition with continuous ethnicity  (2024), arxiv

\bibitem{DBLP:journals/corr/abs-2404-04580}
Otroshi{-}Shahreza, H., Ecabert, C., George, A., Unnervik, A., Marcel, S., Domenico, N.D., Borghi, G., Maltoni, D., Boutros, F., Vogel, J., Damer, N., S{\'{a}}nchez{-}P{\'{e}}rez, {\'{A}}., Mas{-}Candela, E., Calvo{-}Zaragoza, J., Biesseck, B., Vidal, P., Granada, R., Menotti, D., DeAndres{-}Tame, I., Cava, S.M.L., Concas, S., Melzi, P., Tolosana, R., Vera{-}Rodr{\'{\i}}guez, R., Perelli, G., Orr{\`{u}}, G., Marcialis, G.L., Fi{\'{e}}rrez, J.: {SDFR:} synthetic data for face recognition competition. CoRR  \textbf{abs/2404.04580} (2024)

\bibitem{DBLP:conf/icb/OtroshiShahrezaGM23}
Otroshi{-}Shahreza, H., George, A., Marcel, S.: Synthdistill: Face recognition with knowledge distillation from synthetic data. In: {IEEE} International Joint Conference on Biometrics, {IJCB} 2023, Ljubljana, Slovenia, September 25-28, 2023. pp. 1--10. {IEEE} (2023). \doi{10.1109/IJCB57857.2023.10448642}, \url{https://doi.org/10.1109/IJCB57857.2023.10448642}

\bibitem{DBLP:conf/wacv/SenguptaCCPCJ16}
Sengupta, S., Chen, J., Castillo, C.D., Patel, V.M., Chellappa, R., Jacobs, D.W.: Frontal to profile face verification in the wild. In: {WACV}. pp.~1--9. {IEEE} Computer Society (2016)

\bibitem{DBLP:conf/iccv/WangDHTH19}
Wang, M., Deng, W., Hu, J., Tao, X., Huang, Y.: Racial faces in the wild: Reducing racial bias by information maximization adaptation network. In: {ICCV}. pp. 692--702. {IEEE} (2019)

\bibitem{DBLP:journals/pami/WangZD22}
Wang, M., Zhang, Y., Deng, W.: Meta balanced network for fair face recognition. {IEEE} Trans. Pattern Anal. Mach. Intell.  \textbf{44}(11),  8433--8448 (2022)

\bibitem{zheng2018cross}
Zheng, T., Deng, W.: Cross-pose lfw: A database for studying cross-pose face recognition in unconstrained environments. Beijing University of Posts and Telecommunications, Tech. Rep  \textbf{5}(7), ~5 (2018)

\bibitem{DBLP:journals/corr/abs-1708-08197}
Zheng, T., Deng, W., Hu, J.: Cross-age {LFW:} {A} database for studying cross-age face recognition in unconstrained environments. CoRR  \textbf{abs/1708.08197} (2017)

\end{thebibliography}
\end{document}